\documentclass[letterpaper]{article} 
\usepackage{aaai25}  
\usepackage{times}  
\usepackage{helvet}  
\usepackage{courier}  
\usepackage[hyphens]{url}  
\usepackage{graphicx} 
\usepackage{amsmath}
\usepackage{makecell}
\usepackage{booktabs} 
\usepackage{array}    
\usepackage{multirow} 

\usepackage{natbib}  
\usepackage{caption} 
\usepackage{algorithm}
\usepackage{algorithmic}

\usepackage{newfloat}
\usepackage{listings}
\DeclareCaptionStyle{ruled}{labelfont=normalfont,labelsep=colon,strut=off} 
\floatstyle{ruled}
\newfloat{listing}{tb}{lst}{}
\floatname{listing}{Listing}
\title{LLM-PCGC: Large Language Model-based Point Cloud Geometry Compression}
\author{
    Yuqi Ye\textsuperscript{\rm 1}, 
    Wei Gao\textsuperscript{\rm 1}\thanks{Corresponding author. Under review.}
}
\affiliations{
    \textsuperscript{\rm 1}School of Electronic and Computer Engineering, Peking University \\
    yeyuqi0303@stu.pku.edu.cn, gaowei262@pku.edu.cn
}
\usepackage{bibentry}

\begin{document}

\maketitle

\begin{abstract}
The key to effective point cloud compression is to obtain a robust context model consistent with complex 3D data structures. Recently, the advancement of large language models (LLMs) has highlighted their capabilities not only as powerful generators for in-context learning and generation but also as effective compressors. These dual attributes of LLMs make them particularly well-suited to meet the demands of data compression. Therefore, this paper explores the potential of using LLM for compression tasks, focusing on lossless point cloud geometry compression (PCGC) experiments. However, applying LLM directly to PCGC tasks presents some significant challenges, i.e., LLM does not understand the structure of the point cloud well, and it is a difficult task to fill the gap between text and point cloud through text description, especially for large complicated and small shapeless point clouds. To address these problems, we introduce a novel architecture, namely the Large Language Model-based Point Cloud Geometry Compression (LLM-PCGC) method, using LLM to compress point cloud geometry information without any text description or aligning operation. By utilizing different adaptation techniques for cross-modality representation alignment and semantic consistency, including clustering, K-tree, token mapping invariance, and Low Rank Adaptation (LoRA), the proposed method can translate LLM to a compressor/generator for point cloud. To the best of our knowledge, this is the first structure to employ LLM as a compressor for point cloud data. Experiments demonstrate that the LLM-PCGC outperforms the other existing methods significantly, by achieving -40.213\% bit rate reduction compared to the reference software of MPEG Geometry-based Point Cloud Compression (G-PCC) standard, and by achieving -2.267\% bit rate reduction compared to the state-of-the-art learning-based method. 
\end{abstract}

%

\begin{figure}
  \centering
  \includegraphics[width=\linewidth]{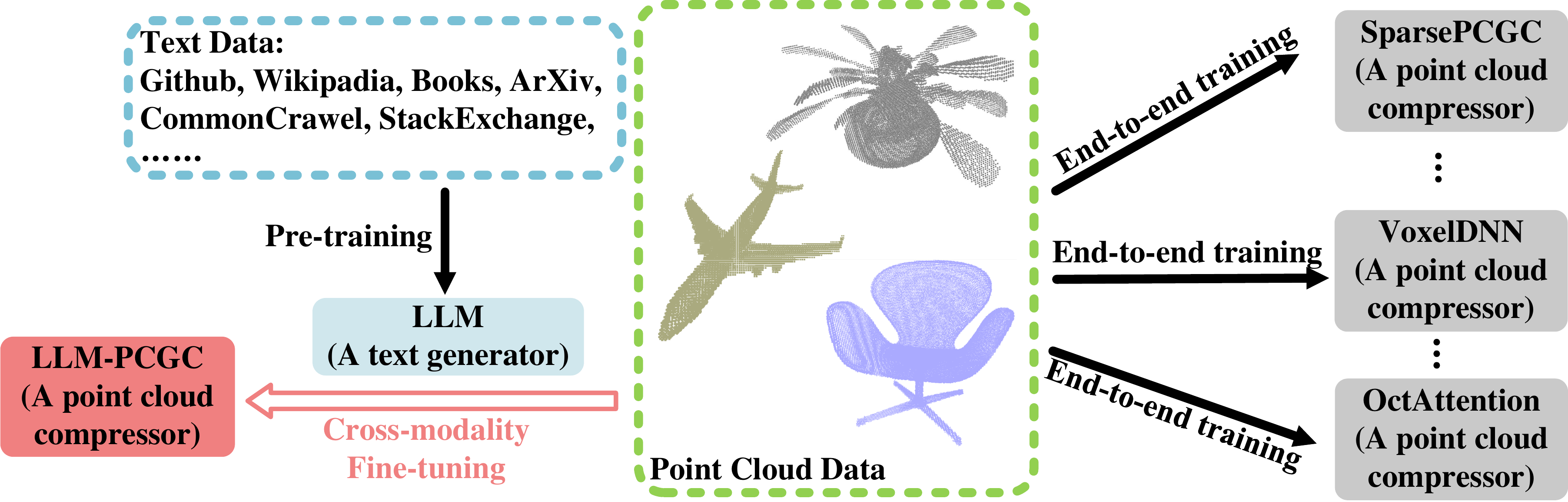}
  \caption{Comparison of training schemes between the proposed LLM-PCGC method and other learning-based methods for point cloud geometry compression. Different from existing methods adopting the end-to-end training manner, our method implements a point cloud compressor by fine-tuning a pre-trained text generator LLM to achieve efficient cross-modality representation alignment. }
  \label{fig:generator_to_compressor}
\end{figure}

\section{Introduction}
Point cloud is a critical and valuable data structure for autonomous driving and virtual reality. Recently, with the development of deep neural networks, more and more learning-based architectures for the lossless PCGC task \cite{que2021voxelcontext, fu2022octattention, wang2022sparse} are proposed, which have demonstrated remarkable performance in the task of lossless PCGC. For these methods, they can be divided into two main categories, i.e., voxel-based and tree-based. Whether voxel-based or tree-based methods, the key to compression performance is the establishment of a strong and robust context model. However, the context capabilities of previous methods remain significantly restricted due to the limitations in data volume and model size, as discussed in the scaling law for large language models (LLMs) \cite{kaplan2020scaling}. This inspires us to directly replace the original context model with LLM, which has large-scale context and generation capabilities.

\begin{figure*}[] 
  \centering
  \includegraphics[width=\textwidth]{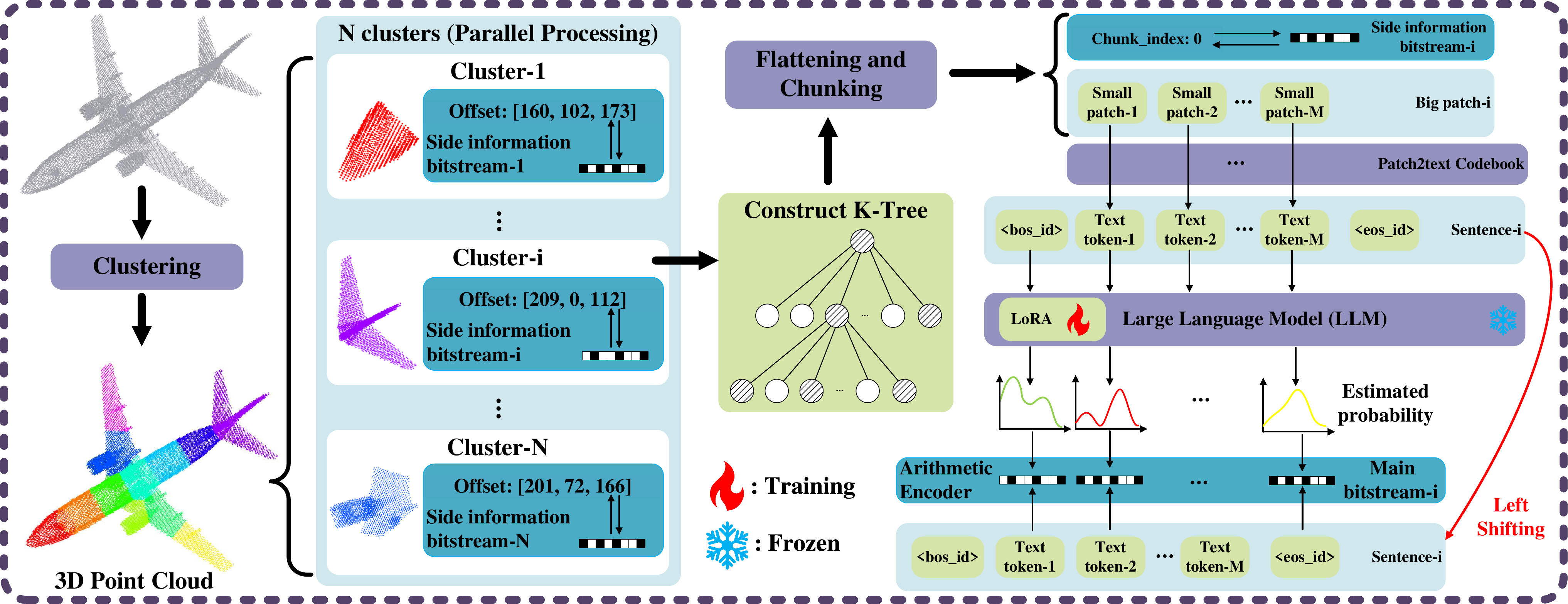}
  \caption{LLM-PCGC encoding pipline. Given a 3D point cloud, the encoding pipeline starts with clustering, followed by normalization and K-Tree structuring. It then employs token mapping invariance for token conversion. Subsequently, a trained LoRA model with a frozen LLM is used to compute the probability distribution for the next token. These distribution are then fed into an arithmetic encoder, resulting in the generation of the encoded bitstream.}
  \label{fig:LLM-PCGC encoding pipline}
\end{figure*}

The emerging viewpoint illustrates that the essence of LLMs fundamentally lies in their ability to compress information \cite{deletang2024language, Li2024EvaluatingLL, valmeekam2023llmzip, yu2024white}. However, prior research works only generally discuss the compact feature representation in LLM from the perspective of compression, but ignore the potential of data compression with LLM. Although in \cite{deletang2024language}, the discussion centers on the lossless compression capabilities of a text-only trained LLM across different modalities of 1D and 2D data, including text, image, and speech. Two limitations emerge from the analysis: 1) This work neglects the compression problem for 3D point clouds. Different from the simple 1D and 2D data, 3D structural data requires a more elaborated and powerful context model. Therefore, as a more complex data type, point cloud owns unique 3D structural characteristics, leading to new challenges for compression task. 2) The exploration of the LLM's data compression capability is restricted to in-context learning, without any additional parameter training. This shows the inherent data compression potential of LLM model, while the performance improvement is still unknown after tailore training for the compression task. In this paper, we propose a completely new architecture, namely the Large Language Model-based Point Cloud Geometry Compression (LLM-PCGC) method, which can better adapt to the lossless PCGC compression task. 

Converting text-based LLM to LLM-PCGC is a cross-modal problem. Since LLM is a model based on text, the current multi-modal large language model (MM-LLM) in order to process multi-modal data, the unified approach is to map other modal tokens to the text space and then generate the modal data through text description \cite{yin2023shapegpt, hong20233d, xu2023pointllm}. On the one hand, for coding tasks, we do not really need text data, and there is no text data to pair with multimodality. On the other hand, we utilize LLMs for their potent generative and contextual understanding capabilities, yet for encoding tasks, the text-based features are extraneous. Hence, we seek to discard the text-specific aspects while maintaining the essential generative and contextual functions. Inspired by \cite{generalpatternmachines2023}, we are the first to fine-tune the pre-trained LLM to achieve cross-modality via token mapping invariance.

Through the above methods, we propose large language model-based point cloud geometry compression (LLM-PCGC). As shown in Fig. \ref{fig:generator_to_compressor}, a comparison is made with existing end-to-end deep learning training methods. Our approach, LLM-PCGC, fine-tunes a pre-trained text generator LLM with point cloud data, achieving cross-modality and serving as a point cloud compressor. In the encoding phase, the procedure begins with the clustering of the input 3D point clouds. Subsequently, each cluster is processed in parallel through a series of steps. First, the coordinates are normalized by subtracting an offset, and a K-tree structure is organized to systematize the point cloud data. Then, the hierarchical tree structure is flattened and divided into segments. Subsequently, a codebook is utilized to translate the point cloud tokens into text tokens to construct an analogous linguistic sentence. Finally, a trained LoRA architecture is employed with a frozen LLM to predict the probability distribution of the next token, which is integrated with an arithmetic encoder to complete the encoding process. The decoding phase mirrors the aforementioned steps in reverse order, thereby reconstructing the original point cloud geometry from the encoded data. 

\begin{figure*}[] 
  \centering
  \includegraphics[width=\textwidth]{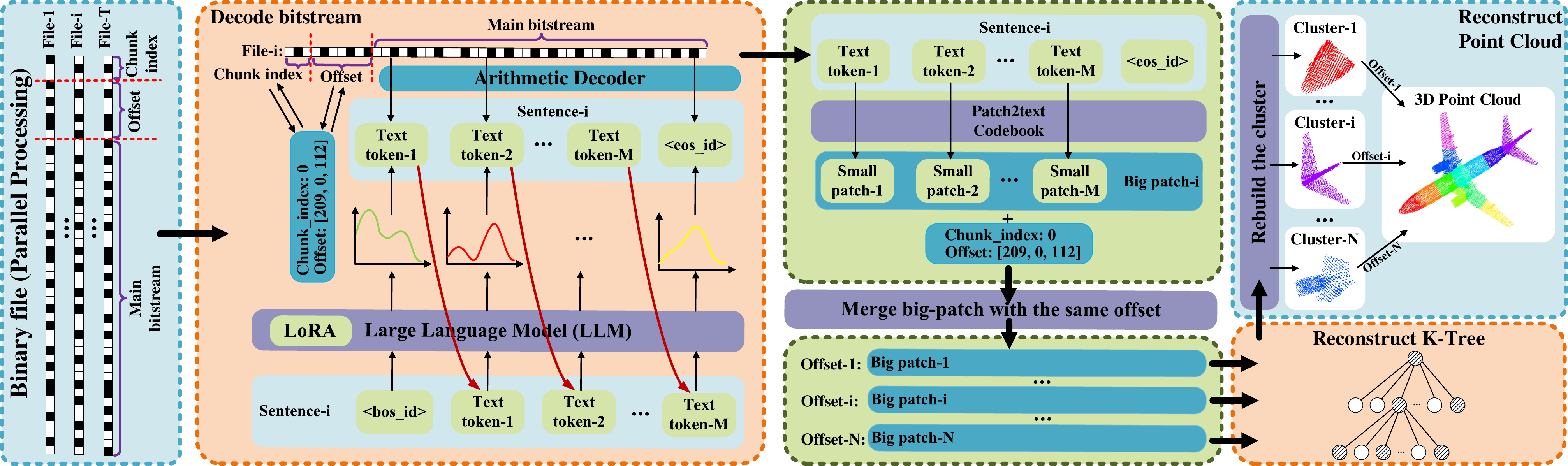}
  \caption{LLM-PCGC decoding pipline. In decoding, binary bits are split, converted to decimals, and the main bitstream is processed in parallel. Through arithmetic decoder, bitstream is decoded by probabilities using LoRA and LLM, and then further mapped into point cloud patches. These patches are aligned and merged by offsets and indices. In the final decoding phase, big patches are structured into a K-Tree for clustered point cloud reconstruction. In the final post-reconstruction, offsets are applied to rebuild the original point cloud.}
  \label{fig:LLM-PCGC decoding pipline}
\end{figure*}

Our contributions are summarized as follows:
\begin{itemize}
    \item We propose a novel architecture, namely LLM-PCGC, which
    is the first to apply LLM as a compressor to point cloud compression within the ``Generator is compressor'' framework. To the best of our knowledge, LLM-PCGC is also the first to transform LLM to a large model that can understand point cloud structure without any text information assistance.
    \item We propose utilize different adaptation techniques for cross-modality representation alignment and semantic consistency, including clustering, K-tree, token mapping invariance, and LoRA, the proposed method can translate LLM to a compressor/generator for point cloud. The approach of token mapping invariance can be transferred to other modalities, offering a new paradigm for multimodal and cross-modal applications of LLMs.
    \item Experiments demonstrate that the LLM-PCGC outperforms the other existing methods significantly, by achieving -40.213\% bit rate reduction compared to the G-PCC, and by achieving -2.267\% bit rate reduction compared to the state-of-the-art learning-based method. As the first LLM-based point compression method, the proposed LLM-PCGC method achieves superior performances.
\end{itemize}

\section{Framework Overview}
\subsection{Encoding Pipline} The LLM-PCGC encoding pipeline is depicted in Fig. \ref{fig:LLM-PCGC encoding pipline}. The encoding phase initiates with the clustering of input 3D point clouds, where each cluster undergoes parallel processing. This process involves several key steps, including normalization of coordinates through offset subtraction, organization of data using a K-tree structure, flattening and chunking of the hierarchical tree structure, and translation of the point cloud's patch tokens into text tokens with a codebook. Special tokens, such as $<$bos\_id$>$ to denote the beginning token id, and $<$eos\_id$>$ to denote the end token id, are incorporated to construct analogous linguistic sentences. The encoding process culminates with the employment of a trained LoRA architecture in conjunction with a frozen LLM to predict the next token's probability distribution, which is then encoded using an arithmetic encoder.

\subsection{Decoding Pipline} The LLM-PCGC decoding pipeline is depicted in Fig. \ref{fig:LLM-PCGC decoding pipline}. In the decoding phase, the parallel-received binary files are segmented to identify the corresponding offset, $<$chunk\_index$>$, and the main bitstream for each bitstream. These identifiers are converted from binary to decimal values, facilitating the processing of the main bitstream through a trainable LoRA and a frozen LLM model to obtain a probability distribution for the next token, which is decoded using an arithmetic coder. The codebook then translates text tokens back into point cloud patch tokens, which are aligned and merged based on their common offset and chunk indexes. An algorithm is applied to reconstruct trees and coordinates. This algorithm ingeniously restores coordinates by counting the number of one, due to the lack of ancestral information. Finally, the cluster point clouds are adjusted by their respective offsets, reconstructing the full point cloud to its original form.

\begin{table*}[t]
\centering
\caption{Bpp performance gains compared to G-PCC and SparsePCGC anchors on MPEG 8i and Owlii datasets.}
\label{tab:performance_SparsePCGC}
\begin{tabular}{@{}llccccc@{}}
\toprule
\textbf{Dataset} & \textbf{Frame} & \textbf{\begin{tabular}[c]{@{}c@{}}G-PCC\\ v14\end{tabular}} & \textbf{\begin{tabular}[c]{@{}c@{}}SparsePCGC\\ 8-stage\end{tabular}} & \textbf{\begin{tabular}[c]{@{}c@{}}LLM-PCGC\\ (Ours)\end{tabular}} & \textbf{\begin{tabular}[c]{@{}c@{}}Gain over\\ G-PCC\end{tabular}} & \textbf{\begin{tabular}[c]{@{}c@{}}Gain over\\ SparsePCGC\end{tabular}} \\
\midrule
\multirow{4}{*}{MPEG 8i} & Longdress\_vox10\_1300 & 1.015 & 0.643 & \textbf{0.631} & -37.882\% & -1.944\% \\
                         & Redandblack\_vox10\_1550 & 1.100 & 0.714 & \textbf{0.703} & -36.100\% & -1.555\% \\
                         & Soldier\_vox10\_0690 & 1.013 & 0.653 & \textbf{0.634} & -38.456\% & -2.925\% \\
                         & Loot\_vox10\_1200 & 0.970 & 0.614 & \textbf{0.597} & -38.454\% & -2.769\% \\
\midrule
\multirow{2}{*}{Owlii} & Basketball\_player\_vox11\_0200 & 0.898 & 0.497 & \textbf{0.490} & -45.479\% & -1.410\% \\
                       & Dancer\_vox11\_0001 & 0.880 & 0.500 & \textbf{0.485} & -44.909\% & -3.079\% \\
\midrule
\textbf{Average} & & 0.982 & 0.603 & \textbf{0.590} & -40.213\% & -2.267\% \\
\bottomrule
\end{tabular}
\end{table*}

\section{Experiments}
\subsection{Training and Testing Setting}
\subsubsection{Data Processing}
Given the current methods which autoregressive methods like OctAttention \cite{fu2022octattention}, VoxelDNN \cite{nguyen2021lossless}, MSVoxelDNN \cite{nguyen2021multiscale}, and NNOC \cite{kaya2021neural} utilize Microsoft Voxelized Upper Bodies (MVUB) \cite{loop2016microsoft} and 8i Voxelized Full Bodies (MPEG 8i) \cite{d20178i} datasets for training, and other approaches like SparsePCGC \cite{wang2022sparse} are trained on ShapeNet \cite{chang2015shapenet}, we aim to ensure fair comparison by training two sets of LLM-PCGC parameters on similar datasets. 

In our experimental comparison with SparsePCGC, we group ModelNet40 \cite{wu20153d} point cloud data into 12 clusters, then organize 3D point cloud clusters in a K-Tree structure with K=12 for training. For testing, we follow the common test condition (CTC) \cite{schwarz2018common}, which recommend to evaluate two public datasets, i.e., MPEG 8i and Owlii \cite{xu2017owlii}.

In relation to autoregressive methods such as OctAttention, VoxelDNN, MSVoxelDNN, and NNOC, we align with the norm by employing widely-used sequences for training. Specifically, we use the point cloud sequences of Andrew10, David10, and Sarah10 from the MVUB, as well as the point cloud sequences of Longdress10 and Soldier10 from MPEG 8i for training. 
 We do a similar clustering process with the cluster number of 240 and K-Tree structure with K=12 for the chosen data. For testing, we select two point clouds from MPEG 8i, Thaidancer and Boxer, which both are downsampled from 12-bit to 10-bit resolution.

\begin{table}[t]
\centering
  \caption{Trainable parameters of LLaMA2-7B \cite{touvron2023llama} used in the proposed LLM-PCGC.}
  \label{tab:Trainable Parameters}
  \setlength{\tabcolsep}{3pt} 
  \begin{tabular}{lcc}
    \toprule
    Configuration & LoRA & Embedding \\
    \midrule
    Component & \makecell{[q, v, k, o, gate, up, \\down] in all layers} & \makecell{[lm\_head, \\ embed\_tokens]} \\
    Hyperparameters & $r=64$, $\alpha=128$ & ------ \\
    Trainable params & 159,907,840 & 295,698,432 \\
    LLaMA2-7B params & 6,738,415,616 & 6,738,415,616 \\
    Trainable\% & 2.37\% & 4.39\%\\
    \bottomrule
  \end{tabular}
\end{table}

\subsubsection{Base Model and LoRA Setting}
Based on LLaMA \cite{touvron2023llama}, an open-source LLM that competes in performance with GPT-3 \cite{brown2020language}, and taking into consideration the hardware resources available, we choose the smallest model, LLaMA2-7B, as our foundational model for this experiment.  As delineated in Table \ref{tab:Trainable Parameters}, we provide a detailed description of the LoRA and Embedding modules of LLaMA2-7B. The total number of trainable parameters amounts to only 6.7\% of the original base LLaMA2-7B model's parameters.
Our model is developed in PyTorch and runs on a system with Intel Xeon Gold 6248R CPUs and only an NVIDIA A40 GPU with 48GB of memory.

\begin{figure}
  \centering
  \includegraphics[width=\linewidth]{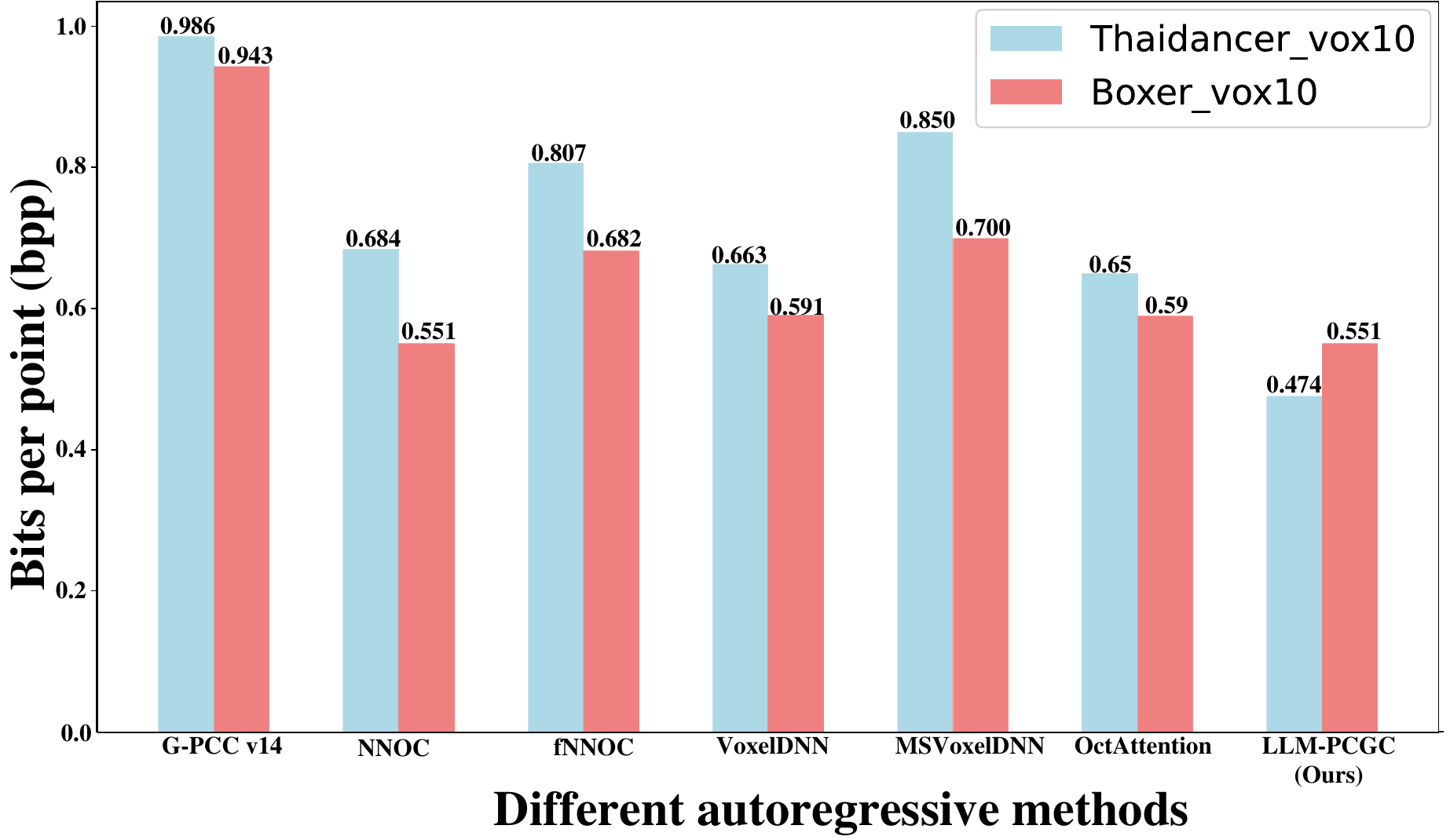}
  \caption{Comparison of bpp among autoregressive methods and traditional method G-PCC.}
  \label{fig:performance_autoregressive}
\end{figure}

\subsection{Experiment Results}
\label{sec:compression_performance}
For experimental comparison with SparsePCGC, we attempt to reproduce SparsePCGC using the similar synthetic ModelNet40 dataset, with bits per point (bpp) performance results presented in Table \ref{tab:performance_SparsePCGC}. For MPEG 8i and Owlii dataset, our proposed LLM-PCGC method achieves -40.213\% bit rate reduction compared to the reference software G-PCC v14 on average and up to 44.909\% for Dancer\_vox11\_0001 and 45.479\% for Basketball\_player\_vox11\_0200. Our method also outperforms achieves -2.267\% bit rate reduction compared to the state-of-the-art learning-based SparsePCGC method.

Regrettably, for the other autoregressive methods, we are not able to reproduce their results by ourselves due to the lack of source codes and relevant materials. Consequently, we reference the performance metrics directly as reported in their original publications. It should be noted that the Thaidancer\_vox10 and Boxer\_vox10 datasets serve as the shared test sets for the evaluation of other autoregressive methods. Accordingly, our examination is confined to the evaluation of compression efficacy on these two specific point cloud datasets, as shown in Fig. \ref{fig:performance_autoregressive}. Building upon the identical G-PCC benchmark, the proposed LLM-PCGC achieves the lowest bpp rate. For instance, it records an average of 0.52 bpp for the MPEG 8i dataset. This marks a reduction of 0.20 bpp from the results achieved by both MSVoxelDNN and fNNOC. In contrast to the OctAttention, which necessitates the inclusion of 1024 neighboring nodes for context modeling, our LLM-PCGC leverages a more robust context model. Remarkably, even in the absence of assistance from ancestor nodes, it still manages to achieve a reduction of 0.10 bpp compared to OctAttention. 

\section{Conclusion}
In this paper, we propose the LLM-PCGC method, which is the first to employ LLM as compressor for the point cloud compression task within the ``Generator is compressor'' framework. We utilize different adaptation techniques, i.e., clustering, K-tree, token mapping invariance, and LoRA, to achieve efficient cross-modality representation alignment and semantic consistency. Without any text data, a text generator can be translated to a point cloud compressor. Experimental results show that the proposed LLM-PCGC method achieves superior compression performance over G-PCC and the state-of-the-art learning-based method, demonstrating the potential of LLMs in data compression. Although as the first attempt to develop a LLM-based point compression method, the proposed LLM-PCGC method achieves superior performances, future research efforts can be made for optimizing the issues on the excessive memory consumption and the long inference time of LLMs.

\bibliography{aaai25}

\end{document}